 \documentclass[smallabstract,smallcaptions]{dccpaper}

\usepackage{epsfig}
\usepackage{citesort}
\usepackage{amsmath}
\usepackage{amssymb}
\usepackage{color}
\usepackage{url}
\usepackage{multirow}
\usepackage{indentfirst}
\usepackage{makecell}
\newlength{\figurewidth}
\newlength{\smallfigurewidth}
\usepackage{graphicx}
\usepackage{subfigure}
\usepackage{fancyhdr}
\usepackage{booktabs}

\begin{document}

\title
{\large
\textbf{Progressive Feature Fusion Network for Enhancing Image Quality Assessment}
}

\author{%
Kaiqun Wu, Xiaoling Jiang, Rui Yu$^{\dag}$, Yonggang Luo, Tian Jiang,\\
Xi Wu, Peng Wei \\
{\small\begin{minipage}{\linewidth}\begin{center}
\begin{tabular}{ccc}
Chongqing Changan Technology Co., Ltd. \\
Chongqing, 401120, China \\
\{wukq, jiangxl, yurui721, luoyg3, jiangtian, wuxi, weipeng4\}@changan.com.cn
\thanks{All authors are with Chongqing Changan Technology Co., Ltd.}
\thanks{$^{\dag}$ Corresponding to Rui Yu (yurui721@changan.com.cn)}
\end{tabular}
\end{center}\end{minipage}}
}

\maketitle
\thispagestyle{empty}

\begin{abstract}
Image compression has been applied in the fields of image storage and video broadcasting. However, it's formidably tough to distinguish the subtle quality differences between those distorted images generated by different algorithms. In this paper, we propose a new image quality assessment framework to decide which image is better in an image group. To capture the subtle differences, a fine-grained network is adopted to acquire multi-scale features. Subsequently, we design a cross subtract block for separating and gathering the information within positive and negative image pairs. Enabling image comparison in feature space. After that, a progressive feature fusion block is designed, which fuses multi-scale features in a novel progressive way. Hierarchical spatial 2D features can thus be processed gradually.  Experimental results show that compared with the current mainstream image quality assessment methods, the proposed network can achieve more accurate image quality assessment and ranks second in the benchmark of CLIC in the image perceptual model track.
\end{abstract}

\Section{1.Introduction}

\indent Advancements in digital technology have significantly increased the complexity and scope of tasks such as image compression\cite{image_compress}\cite{iqa}. For better visual quality of compression, the compressed image needs to be closer to the uncompressed one. However, as the technology of image compression improves, it is difficult to tackle the differences between the compressed/original image pairs. Therefore, it is necessary to create image quality assessment (IQA) methods to measure the compression quality. 

\indent In the field of image compression, people usually evaluate the quality of a compressed image based on the reference image and the distorted image. Some traditional methods such as PSNR\cite{psnr}, SSIM\cite{SSIM}, and MS-SSIM\cite{MS-SSIM} are proposed and have been widely utilized. These traditional methods mainly consider pixel differences and structural differences, while having low computational complexity. However, these methods possess certain limitations in recognizing some edge information and in distinguishing noise resembling authentic textures.

\indent Recent research on IQA concentrates on deep learning-based methods. For instance, IQA-TMFM\cite{IQA-TMFM} employs a multi-index fusion method based on transformer\cite{transformer}, which can extract the feature of the whole image. FFDN\cite{FFDN} differentiates more effectively between distorted and reference feature maps, resulting in better image quality assessment. Furthermore, SwinIQA\cite{swiniqa} offers a full-reference IQA metric for evaluating the perceptual quality of compressed images through a learned swin distance space. However, these methods concatenate the multi-scale features after processing, and the correlation between the features is somehow ignored.

\begin{figure}[t]
    \centering
    \includegraphics[width=0.75\linewidth]{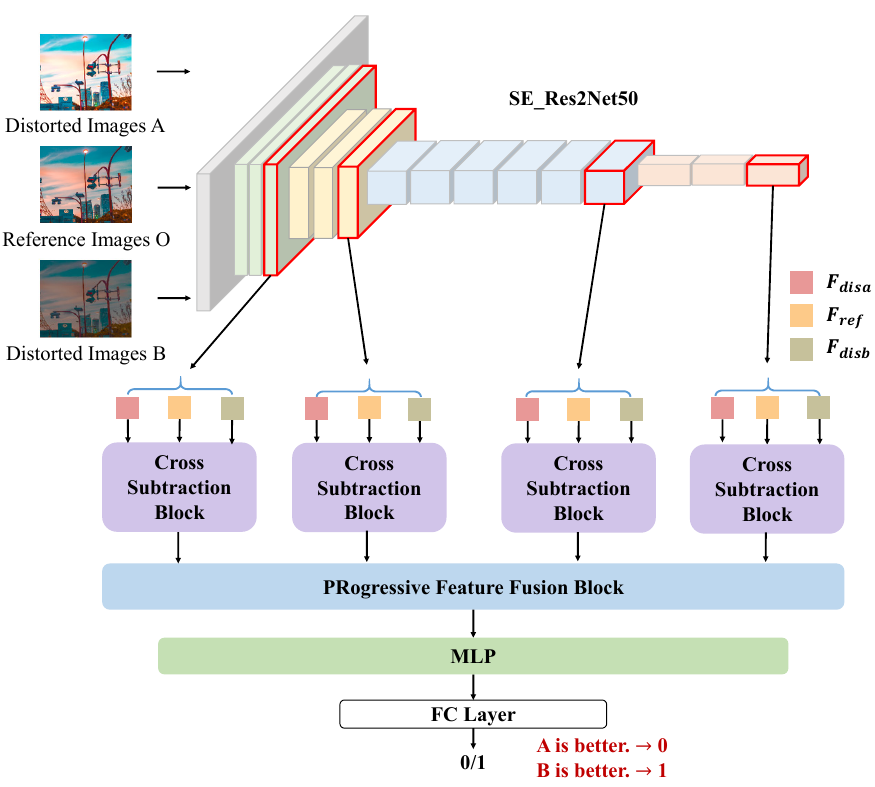}
    \vspace{-5pt}
    \caption{PRFNet Architecture. The input of the network includes the reference image $O$ and two distorted images $A$ and $B$ with different degrees of compression. In the feature extraction module, feature maps of the three images at four scales ($F_{disa}^i$,$F_{ref}^i$,$F_{disb}^i$, where $i= 1, 2, 3, 4$) can be obtained respectively. Instead of measuring the differences between the compressed and reference image in low level space, cross subtraction blocks utilize the differences in the features. A progressive feature fusion block is set to fuse different scales of features progressively. Finally, multi-layer perceptron (MLP) networks are used to get the classification result.}
    \label{fig:architecture}
\end{figure}

\indent In this paper, we propose a novel image quality assessment network named PRogressive feature Fusion Net(PRFNet) which is an end-to-end network to deal with the CLIC image quality assessment challenge. PRFNet mainly consists of three modules, a feature extraction module, a cross subtraction module and a progressive feature module. In the first module, a pre-trained model is leveraged to extract multi-scale features of the images. The attention mechanism is also used to enhance the feature representation. Then, the differences between the reference image and distorted image with subtractive operation are computed in the cross subtraction module. To progressively grasp hints of the correlation between the features, a progressive feature module is applied in PRFNet to fuse the features. Last but not least, a progressive training strategy is used for a more efficient training process. Experimental results have shown that our PRFNet achieves competitive accuracy on the CLIC 2022 val set.

\indent The remainder of this paper is organized as follows. In Section.2, the architecture of PRFNet is introduced. In Section.3, experimental results are displayed. Finally, we conclude in Section.4.

\Section{2.Progressive Feature Fusion Network}
\label{sec:progressive feature fusion network}
\indent The structure of our proposed PRFNet is shown in Figure \ref{fig:architecture}. Based on the requirement of the challenging CLIC image quality assessment task, the provided reference image and both two compressed images are grouped as the inputs of the network. Overall, the network consists of a feature extraction module, a cross subtraction module and a progressive feature fusion module which is followed by a Multi-layer perceptron (MLP) layer to obtain the final classification result. The progressive training strategy is also introduced in this section.

\begin{figure}[htbp]
    \centering
    \includegraphics[width=0.95\linewidth]{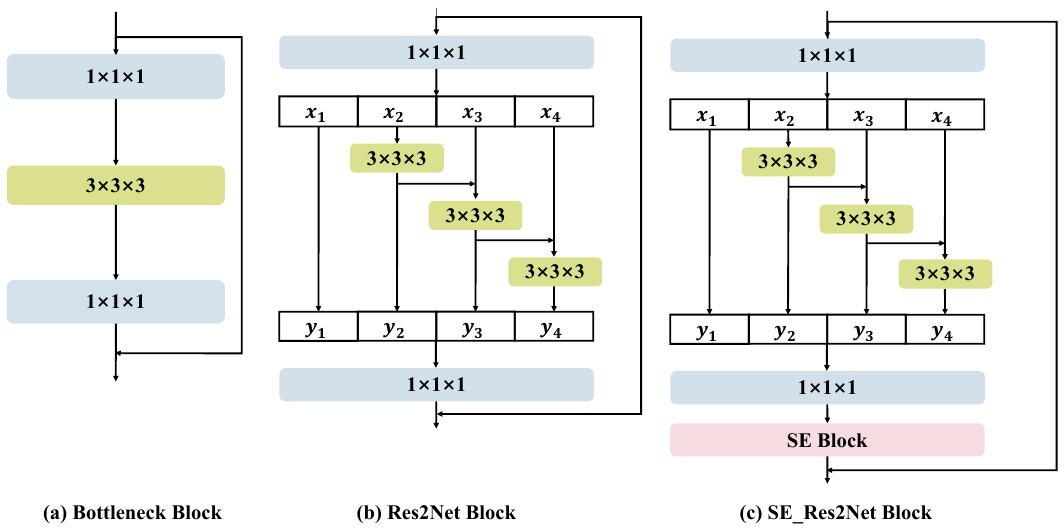}\vspace{-5pt}
    \caption{The blocks of (a)the ResNet bottleneck, (b)the Res2Net block, and (c)the SE\_Res2Net block. The receptive field of a traditional ResNet bottleneck can be relatively narrow, while Res2Net can expand the receptive field by decomposing and recomposing. SE\_Res2Net block can extract more representative features as a SE\_block is followed to grasp the global information.}
    \label{fig:feature_block}
\end{figure}
\vspace{-15pt}
\SubSection{2.1 Feature Extraction Module}
\indent For most of deep learning-based image processing tasks, the very first thing is to extract the feature of the input image. To utilize the image information, the feature extraction module needs to be designed delicately and representatively. For example, different scales of features, including features from shallow layers and deep layers, should be considered, since shallow features concentrate on the pixel-related details of the images, while deep features focus on the semantic information\cite{feature_vis}.

\indent In our feature extraction module, SE\_Res2Net50\cite{res2net} is chosen as the main component. This network is built with four stages of SE\_ResNet blocks, which are modified from the basic ResNet bottlenecks\cite{resnet}. The numbers of SE\_ResNet blocks used in each stage are 3, 4, 6, 3. The minimum downsampled ratio of the features from the first stage is 2. Other settings are following the traditional ResNet50.

\indent As shown in Figure \ref{fig:feature_block}, a bottleneck of the traditional ResNet can only capture features from a narrow receptive field. However, Res2Net blocks can go further, as it decomposes the convolution into multiple submodules and connects these submodules by constructing hierarchical residual connections within a single residual block. The range of the receptive field can be expanded and more abundant features can be extracted. Besides, the structure of Res2Net blocks is compatible with many other feature enhancing blocks. Squeeze-and-excitation block(SE\_block)\cite{seblock} can extract channel attention and learn the global information of the feature. As shown in Figure \ref{fig:feature_block}(c), it can be combined with the original Res2Net blocks to form a SE\_Res2Net block.

\begin{figure}[htbp]
    \centering
    \includegraphics[width=0.85\linewidth]{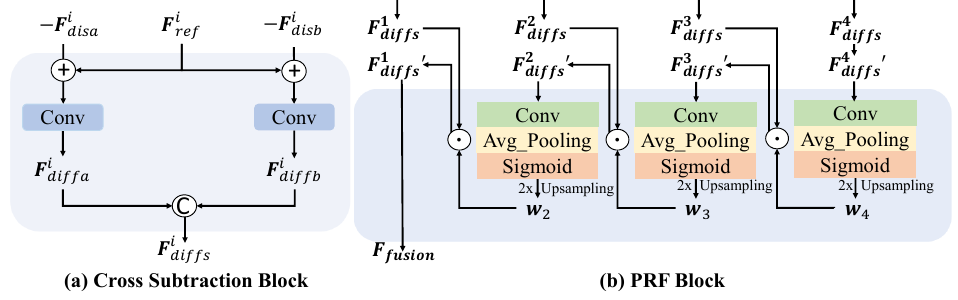}
    \caption{(a)Cross Subtraction Block. This block calculates the differences of feature maps of image $A$, $B$, and $O$, and carries out the feature difference maps by subtraction and cross operations. (b)PRogressive feature Fusion Block. In this block, weights are obtained by a series of operations on deep-level features, then fused with lower-level features by hadamard product.}
    \label{fig:special_block}
\end{figure}
\vspace{-15pt}
\SubSection{2.2 Cross Subtraction Module}
\indent The cross subtraction module contains four cross subtraction blocks. Each block processes a certain scale of features from the previous feature extraction module. As Figure \ref{fig:special_block}(a) shows, two operations are applied in the block. The first one is subtraction, where the network is able to learn the differences between the distorted images and the reference image. The second one is the cross operation, which builds the relationship between the feature differences.

\indent In detail, we denote the features of image $O$, $A$, and $B$ at four scales as $F_{ref}^i$, $F_{disa}^i$ and $F_{disb}^i$, where $i=1,2,3,4$. The feature differences $F_{diffa}$ and $F_{diffb}$ can be calculated as,
\begin{equation}
    \begin{array}{c}
    F_{diffa}^i=\textbf{Conv}(F_{ref}^i-F_{disa}^i), \\ 
    F_{diffb}^i=\textbf{Conv}(F_{ref}^i-F_{disb}^i). \\
    \label{eq:diff}
\end{array}
\vspace{-10pt}
\end{equation}

Then, the cross operation is applied on the difference maps through a concat operation in channel dimension, which can be expressed as,
\begin{align}
    F_{diffs}^i=\textbf{concat}(F_{diffa}^i,F_{diffb}^i).
    \label{eq:diffs}
\end{align}
With the cross subtraction module, feature difference maps of four stages $F_{diffs}^1$, $F_{diffs}^2$, $F_{diffs}^3$, $F_{diffs}^4$ can be obtained in the network.

\SubSection{2.3 Progressive Feature Fusion Module} 
\indent To make full use of feature difference maps, it is necessary to fuse these features. Although FPN\cite{fpn} and BiFPN\cite{bifpn} are commonly used in object detection, these two feature fusion methods do not perform well in our experiments. Inspired by the structure of the self-attention block in transformer\cite{transformer}, we propose a progressive feature fusion module for feature fusion.  In practice, this module is built with a progressive feature
fusion block(PRFBlock) where feature difference maps are fused stage by stage as depicted in Figure \ref{fig:special_block}(b).

\indent In detail, feature difference maps $F_{diffs}^i$ are obtained from the previous cross subtraction module. Starting from the deepest feature difference map $F_{diffs}^4$, the fusion is operated stage by stage as,
\vspace{-5pt}
\begin{equation}
{F_{diffs}^{i}}'=
\begin{cases}
    w_{i+1} \odot F_{diffs}^i, \quad i=1,2,3\\
F_{diffs}^i, \quad i=4
\end{cases},
\label{eq:fusion}
\end{equation}
where $\odot$ represents hadamard product and $w_{i+1}$ is calculated as,
\begin{align}
    w_{i+1} = h(F_{diffs}^{i+1}),\quad i=1,2,3.
    \label{eq:weight}
\end{align}
Here, the function $h(\cdot)$ in Formula(\ref{eq:weight}) represents a group of network operations. These operations contain a convolution, an average pooling in channel dimension, a non-linear function(sigmoid) and a 2x upsampling.

\indent After ${F_{diffs}^{1}}'$ is calculated, it is used as the output of the fusion block. We denote the output feature map as $F_{fusion}$, which will be fed into the following MLP network. The final classification result will obtained through a fully connected layer.

\SubSection{2.4 Progressive Training Strategy}
\indent The challenging CLIC image quality assessment task is often treated as a binary classification task. However, as the distorted image $A$ and $B$ are so similar, it is very difficult to decide which image is closer to the reference image $O$. It is not a good solution to use such a hard score in this task. Instead, here we treat the task as a regression task and a soft score is considered. In detail, the network regresses the probability $q_i$ with higher value meaning selecting image $B$ as the closer image. On the contrary, the lower value of $q_i$ indicates that the network selects image $A$ as the more similar one. 

\indent Considering not all the datasets used in the training process have accurate probability labels, we used a two-step progressive training strategy to train our proposed PRFNet. The first training step is the coarse training. In this step, we treat the task as a classification task. The loss is designed as the cross-entropy loss, which is shown in formula (\ref{eq:first_train}) as,
\vspace{-10pt}
\begin{align}
    \mathcal{L}_{BCE} = -\dfrac{1}{N} \sum_{i=1}^{N} ( y_i log( p_i) + (1-y_i) log(1-p_i)),
    \label{eq:first_train}
\end{align}
where $p_i$, $y_i$ and $N$ represents of the probability the $i-th$ sample to select image $B$ as the closer image to image $O$, the binary label to select image $B$ as the closer image to image $O$, and the number of samples respectively. The second training step is the fine training. In the second step, the image quality assessment task is viewed as a regression task. The loss function is designed as the regression task, which is shown in formula (\ref{eq:second_train}) as,
\vspace{-10pt}
\begin{align}
    \mathcal{L}_{MSE}=\dfrac{1}{N} \sum_{i=1}^{N} ( q_i - \hat{q_i} ) ^ 2,
    \label{eq:second_train}
\end{align}

where $q_i$, $\hat{q_i}$ and $N$ refer to the predicted score, the soft label and the number of samples respectively.

\indent It should be noted that both types of tasks can be trained in a multi-task learning (MTL) manner. Although multi-task learning is used in many image processing tasks\cite{mtl} and may simplify the training procedure. However, for image quality assessment tasks, multi-task learning may not be proper as shown in Section 3.3.2.

\Section{3.Experiments}
\label{sec:experiments}
\SubSection{3.1 Datasets}
\indent In our work, three datasets are used in the training process and one dataset for validation.\\ 
\indent \textbf{CLIC-T:}This training dataset is provided by the CLIC2021 competition, which contains $122107$ pairs of images and most of them are generated by the compressed methods of traditional codec such as HEVC/H.265\cite{sullivan2012overview} and VVC/H.266\cite{bross2021overview}) and learning-based methods\cite{gao2021perceptual}. Each pair of images contains a reference image $O$, two distorted images $A$ and $B$. The label is $1$ or $0$. $0$ means image $A$ is more similar to image $O$.\\
\indent \textbf{BAPPS-T\cite{bapps}:}This training dataset is a two-alternative forced choice (2AFC) part of Berkeley-Adobe Perceptual Patch Similarity dataset, which includes 187744 pairs of images and all of them are generated by traditional distortion (e.g., lightness shift, color shift, uniform white noise) and convolutional neural network algorithms built in a variety of tasks, architectures and losses. The label is a float in the range of $0$ to $1$, which indicates the probability that people prefer to choose distorted image $B$.\\
\indent \textbf{PieApp\cite{pieapp}:}It contains $82080$ pairs of training images. The distorted images are generated by traditional codec methods or CNN methods.\\
\indent \textbf{CILC-V:}It is the only validation set provided by CLIC2022 competition. 5220 image pairs are used to validate the accuracy of our PRFNet.\\
\indent Quantitative information of the above datasets is shown in Table \ref{tab:dataset}.

\begin{table}[htbp]
    \centering
    \renewcommand\arraystretch{1.2}         
    \caption{Quantitative Information of Image Quality Assessment Datasets}
        \vspace{-5pt}
    \setlength{\tabcolsep}{1pt}
    \begin{tabular}{cc | c}    
        \Xhline{1.3pt}\hline
         \textbf{Dataset}&  \textbf{\makecell{Distort Types}}  &  \textbf{\makecell{Number of\\Image Pairs}}\\
         \Xhline{1.3pt}
           \textbf{CLIC-T}     &Codec outputs         &       $122106$        \\
           
          \textbf{BAPPS-T}   & Traditional+CNN         &     $187744$      \\

           \textbf{PieApp}          &  Traditional+CNN        & $82080$   \\

           \textbf{CILC-V}      & Codec outputs         &       $5220$       \\
         \Xhline{1.3pt}
        \end{tabular}
    
    \label{tab:dataset}
\end{table}

\SubSection{3.2 Implementation details}
\indent Our proposed PRFNet is designed based on the Pytorch framework with an NVIDIA A100 GPU. In the training process of both two steps, batch size is set to $16$ and SGD optimizer is used. At the first training step, only the CLIC-T dataset is used and the training period is $40$ epochs. The initial learning rate is 0.001 and is divided by $10$ every $10$ epochs. At the second training step, the weights of the feature extraction module are frozen. The initial learning rate in the second step is set as $0.0001$ and is divided by $10$ every $10$ epochs. BAPPS-T and PieApp datasets are used and the training period is 20 epochs in the second step. Images are cropped to $448 \times 448 \times 3$ randomly in the training process. The CLIC-V dataset is used in the validation process.

\SubSection{3.3 Ablation Study}
\indent Two ablation experiments are conducted to show the effectiveness of our proposed PRFNet.

\SubSection{3.3.1 Ablation Study on Feature Fusion Method}
\indent To further evaluate the effectiveness of PRFBlock, the accuracy of different feature fusion methods including FPN\cite{fpn} and BiFPN\cite{bifpn} is tested on the CLIC-V dataset. As shown in Table \ref{tab:ablation_fuse_method}, our proposed PRFBlock performs better than the others.
\vspace{-15pt}
\begin{table}[htbp]
    \centering
    \renewcommand\arraystretch{1.2}         \tabcolsep=2pt
    \caption{Quantitative Results on Different Feature Fusion Methods}
        \vspace{-5pt}
    \begin{tabular}{c|c}        
        \Xhline{1.3pt}\hline
         \textbf{Feature Fusion Method} &  \textbf{Accuracy}    \\
         \Xhline{1.3pt}
         FPN                            & $0.741$                      \\
         BiFPN                          & $0.772$                      \\
         \textbf{PRFBlock}                  & $\textbf{0.781}$                      \\
         \Xhline{1.3pt}
        \end{tabular}
    \label{tab:ablation_fuse_method}
\end{table}

\SubSection{3.3.2 Ablation Study on Progressive Training Strategy}
\indent In this subsection, a multi-task learning strategy is considered. In this way, $\mathcal{L}_{MSE}$ and $\mathcal{L}_{BCE}$ are used in the network optimization simultaneously in the same training step. As shown in Table \ref{tab:ablation_training_strategy}, the accuracy of the progressive training strategy is better than that of MTL.

\begin{table}[htbp]
    \centering
    \renewcommand\arraystretch{1.2}         \tabcolsep=2pt
    \caption{Quantitative Results on Different Training Strategies}
    \vspace{-5pt}

    \begin{tabular}{c|c}        
        \Xhline{1.3pt}\hline
         \textbf{Training Strategy} &  \textbf{Accuracy}    \\
         \Xhline{1.3pt}
         Multi-task Learning                          & $0.737$                      \\
         \textbf{OURS}                  & $\textbf{0.781}$                      \\
         \Xhline{1.3pt}
        \end{tabular}
    \label{tab:ablation_training_strategy}
\end{table}

\SubSection{3.4 Comparison with other methods}
\indent We compare our proposed PRFNet with the currently popular methods and those of the award-winning teams in CLIC 2022. In CLIC-V, our method achieved the best accuracy among these methods. Experimental results are shown in Table \ref{tab:final_results}.
\vspace{-15pt}
\begin{table}[htbp]
    \centering
    \renewcommand\arraystretch{1.2}           
    \caption{Quantitative Results on the Accuracy of CLIC-V.}
        \vspace{-5pt}
    \begin{tabular}{cc}        
        \Xhline{1.3pt}\hline
         \textbf{Methods}       &\textbf{CLIC-V}  \\
         \Xhline{1.3pt}
         SSIM\cite{SSIM}        &$0.571$\\
         PSNR                   &$0.572$\\
         MS-SSIM\cite{MS-SSIM}  &$0.612$\\
         LPIPS\cite{bapps}      &$0.740$\\
         FFDN\cite{FFDN}        &$0.762$\\
         SwinIQA\cite{swiniqa}  &$0.780$\\
         IQA-TMFM\cite{IQA-TMFM}&$0.780$\\
         alexkkir	           &$0.813$\\
         \Xhline{1.3pt}
         \textbf{PRFNet}          &$\textbf{0.781}$\\
         \Xhline{1.3pt}
        \end{tabular}
    \label{tab:final_results}
\end{table}
\vspace{-5pt}
\Section{4.Conclusion}
\label{sec:conclusion}
\indent In this paper, we propose a full-reference image quality assessment method, PRFNet, to select a more similar compressed image to the reference image. PRFNet is built with a feature extraction module, a cross subtraction module and a progressive feature fusion module. A progressive training strategy is also introduced for better IQA accuracy. Experiments have demonstrated that our approach achieves competitive results on CLIC-V datasets.

\Section{References}
\bibliographystyle{IEEEbib}
\bibliography{refs}

\end{document}